\documentclass[journal]{journal}
 \usepackage{graphicx}
 \usepackage{enumerate}
 \usepackage{mathrsfs}
 \usepackage{utfsym}
 \usepackage{amsmath}
 \usepackage{amssymb}
\usepackage{multirow}

\ifCLASSINFOpdf
\else
\fi
\begin{document}
%
\title{TGP: Two-modal occupancy prediction with 3D Gaussian and sparse points for 3D Environment Awareness}
%
%
%

\author{Mu~Chen,
        Wenyu~Chen,
        Mingchuan~Yang,~\IEEEmembership{Member,~IEEE},
        Yuan~Zhang,
        Tao~Han,~\IEEEmembership{Member,~IEEE},
        Xinchi~Li,
        Yunlong~Li,
        and Huaici~Zhao
        
\thanks{Mu Chen is with the China Telecom Research Institute, Beijing 102200, China, and the University of Chinese Academy of Sciences, Beijing 100049, China (e-mail: chenm72@chinatelecom.cn).}
\thanks{Yunlong Li, Mingchuan Yang, Yuan Zhang, Tao Han, Xinchi Li and Yuan Zhang are with the China Telecom Research Institute, Beijing 102200, China.}
\thanks{Wenyu Chen, Huaici Zhao are with the Key Laboratory of Opto-Electronic Information Processing, Shenyang Institute of Automation, Chinese Academy of Sciences, Shenyang 110016, China, and also with the University of Chinese Academy of Sciences, Beijing 100049, China.}
\thanks{Mu Chen and Wenyu Chen contribute equally to this work.}
\thanks{Corresponding author: Wenyu Chen (e-mail: chenwenyu@sia.cn).}
\thanks{Manuscript received XXX; revised XXX.}}

%
%

\markboth{Journal of \LaTeX\ Class Files,~Vol.~6, No.~1, January~2007}%
{Shell \MakeLowercase{\textit{et al.}}: Bare Demo of IEEEtran.cls for Journals}
%



\maketitle
\thispagestyle{empty}

\begin{abstract}
3D semantic occupancy has rapidly become a research focus in the fields of robotics and autonomous driving environment perception due to its ability to provide more realistic geometric perception and its closer integration with downstream tasks. By performing occupancy prediction of the 3D space in the environment, the ability and robustness of scene understanding can be effectively improved. However, existing occupancy prediction tasks are primarily modeled using voxel or point cloud-based approaches: voxel-based network structures often suffer from the loss of spatial information due to the voxelization process, while point cloud-based methods, although better at retaining spatial location information, face limitations in representing volumetric structural details. To address this issue, we propose a dual-modal prediction method based on 3D Gaussian sets and sparse points, which balances both spatial location and volumetric structural information, achieving higher accuracy in semantic occupancy prediction. Specifically, our method adopts a Transformer-based architecture, taking 3D Gaussian sets, sparse points, and queries as inputs. Through the multi-layer structure of the Transformer, the enhanced queries and 3D Gaussian sets jointly contribute to the semantic occupancy prediction, and an adaptive fusion mechanism integrates the semantic outputs of both modalities to generate the final prediction results. Additionally, to further improve accuracy, we dynamically refine the point cloud at each layer, allowing for more precise location information during occupancy prediction. We conducted experiments on the Occ3D-nuScenes dataset, and the experimental results demonstrate superior performance of the proposed method on IoU-based metrics.
\end{abstract}


%
\IEEEpeerreviewmaketitle

\section{Introduction}
%
%
%
%
\IEEEPARstart{I}{n} 3D perception for robotics and autonomous driving, cubic bounding boxes \cite{bevdepth,BEVFormer,DETR3D,PETR} have been widely used but have limitations in capturing the shape of irregular objects. Additionally, they require extra processing when applied to downstream tasks. On the other hand, 3D occupancy prediction \cite{renderocc,opus,sparseocc,surroundocc} divides the space into grid cells and predicts the occupancy probability for each cell. This approach can better capture objects of any shape and offers a more unified 3D representation for tasks like 4D prediction and path planning. Due to its detailed and adaptable nature, 3D occupancy prediction has gained significant attention and rapid growth in research.

3D occupancy prediction has shifted from focusing on individual objects to enabling full-scene perception \cite{bevdepth,BEVFormer,fbocc,surroundocc}, showing great potential for broader applications. However, this progress comes with significant computational challenges, mainly due to the reliance on dense 3D representations. These dense representations are not essential for occupancy prediction tasks. Voxel-based representations, for example, have fixed resolutions and boundaries, making them inflexible, especially for targets of different sizes and shapes. In sparse scenes, many voxels remain empty, leading to inefficient resource use. To address these issues, several methods \cite{open, sparseocc,opus,tri,voxel} have been proposed. SparseOcc \cite{sparseocc}refines the voxel-based representation progressively, eliminating empty grids and reducing redundant computations. This is particularly efficient in sparse areas. Similarly, OPUS \cite{opus} uses a point-based strategy to replace the uniform 3D grid, offering more flexibility and precision in representing 3D space. This method also adapts better to local features, improving detection accuracy. While both approaches have made progress, voxel-based methods struggle with local information aggregation, and point-based methods face challenges in handling volumetric features.

Inspired by PV-RCNN \cite{pvrcnn} and 3D Gaussian \cite{gaussianformer, 3dGS}, we innovatively proposes a sparse representation method based on the fusion of 3D Gaussians and points. In this approach, points effectively aggregate local information, while 3D Gaussians flexibly capture and represent volumetric information in high-dimensional space, thereby enhancing the ability to represent spatial details. Specifically, we design a dual-modal decoder structure based on the sparse paradigm to progressively refine 3D occupancy predictions. This dual-modal decoder structure combines a Transformer-based framework, taking queries, sparse points, and 3D Gaussian sets as inputs. Such a decoder structure is highly flexible, capable of extracting local features sufficiently at different refinement levels while effectively capturing volumetric features through Gaussian representations. Notably, the position of 3D Gaussians share the same initial values with sparse points. Each 3D Gaussian element includes attributes such as position, volume, rotation, and semantic information. In each layer of the decoder, the initial points are progressively split. These points not only enhance the query features, but also provide positional references for the refinement process in the next layer. Meanwhile, 3D Gaussians interact with the enhanced query features at a fixed number in each layer, further refining the representation of the 3D Gaussians. Through this multi-level interaction process, the model is able to gradually optimize the 3D occupancy prediction and refine the spatial representation at each layer. Based on the semantic predictions from the enhanced queries and refined 3D Gaussian outputs in each layer, we propose an adaptive fusion method to obtain the final occupancy prediction result. 

Overall, our contributions are as follows:
\begin{enumerate}[1)]
\item We propose an innovative dual-modal occupancy prediction decoder structure based on 3D Gaussians and point s, which effectively balances position information and volumetric structure information.
\item Our method achieves good performance on the Occ3D nuScenes dataset.
\end{enumerate} 

\section{Related work}
\subsection{3D Occupancy prediction}
In recent years, image-based detection methods \cite{bevdepth, BEVFormer, DETR3D, PETR} have advanced significantly, improving precision and closing the gap with other sensors. However, traditional methods struggle to capture all the details of a scene in the open world. Vision-based 3D occupancy prediction methods \cite{fbocc, gaussianformer, voxel, tri, surroundocc, sparseocc, opus, renderocc, occformer} offer a promising solution by predicting the spatial occupancy and semantics of 3D voxel grids around autonomous vehicles based on image input, making them a key focus in autonomous driving. Some methods \cite{bevdepth, BEVFormer, occformer} use bird's-eye view (BEV) perception and multi-view representations to enhance spatial perception by improving image features. Others \cite{3dGS, sparseocc, voxel} work directly with 3D voxel representations, which minimize information loss but come with high computational costs. To address this, some methods \cite{open, sparseocc, opus, tri, voxel} explore a coarse-to-fine feature learning approach. 
Voxformer\cite{voxel} uses depth as a prior to predict occupancy and select valuable queries. Only the occupied queries gather information from the image, and the updated queries, along with the masked tokens, collectively reconstruct the voxel features.
SparseOcc \cite{sparseocc} reduces computational demands by prioritizing influential queries. However, these methods often require complex spatial modeling steps. In contrast, OPUS \cite{opus} adopts a point-set prediction paradigm, relying on queries to identify occupied regions, offering more flexible feature aggregation and better positional information. However, point-set-based sparse prediction does not capture the volumetric structure of the scene well. Thus, it is important to design a prediction structure that integrates both volumetric and positional representations.

\subsection{3D Gaussian Splatting}
3D Gaussian splatting (3D-GS) \cite{3dGS} has made significant progress in the field of computer graphics in recent years, particularly in 3D rendering and scene reconstruction tasks\cite{NeRF,ScaffoldGS,DeformableGS}. The 3D-GS technique represents scenes as multiple 3D Gaussian points or small clusters, utilizing these points for rasterization rendering, achieving high-quality rendering effects while maintaining low computational costs\cite{Stopthepop,eagles}. Compared to traditional mesh or voxel representations, 3D-GS is more flexible in representing complex geometries and offers advantages in memory usage and computational efficiency. In autonomous driving, 3D Gaussian has been widely applied in perception, mapping, path planning, and other areas\cite{GaussianAD,GVnet,GSPR}. GaussianFormer \cite{gaussianformer} was the first to introduce 3D Gaussian into occupancy prediction in an online manner. However, the special design mechanism using 3D convolutions to enable interaction between 3D Gaussians results in an increased computational burden.


\section{Methodology}
This section first outlines the network structure of our proposed method and details the key component of the two-modal decoder with Gaussian and point representations as inputs. Finally, we introduce the supervision method used.

\begin{figure*}[t]
\centering
\includegraphics[width=6in]{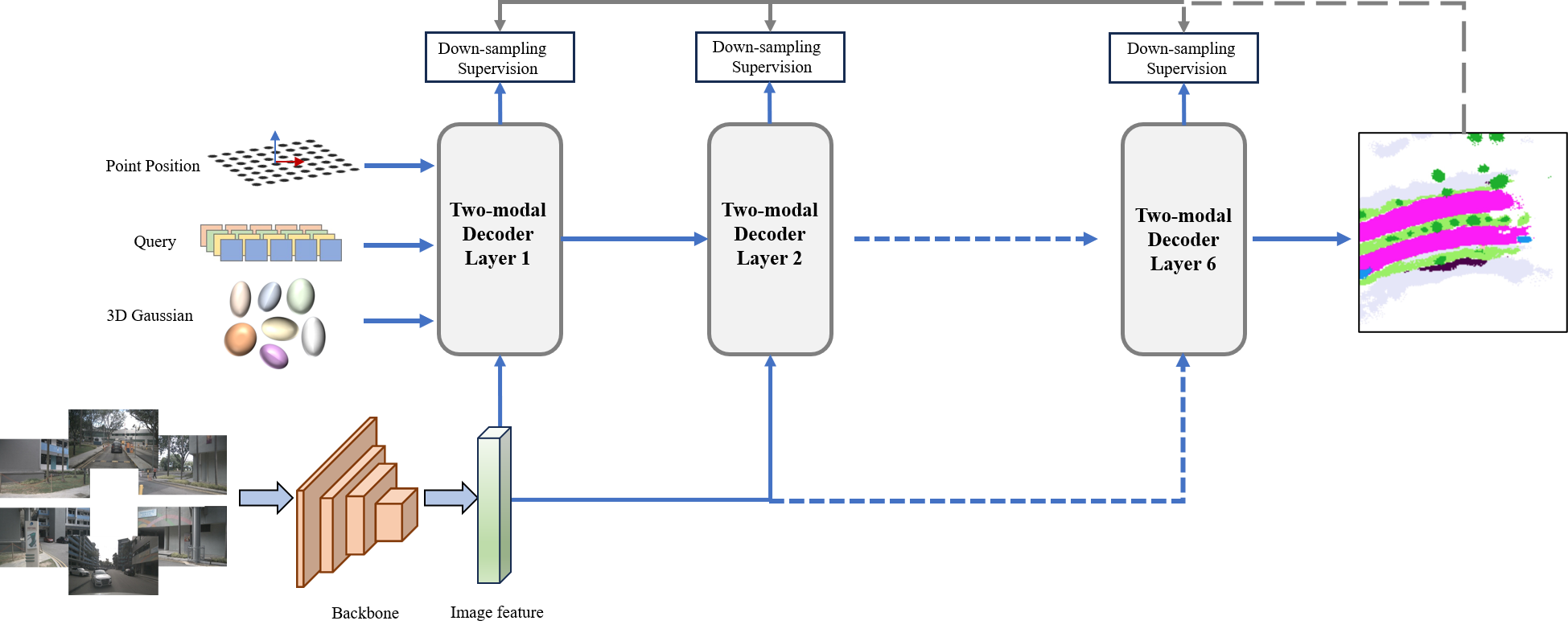}
\caption{Framework of the proposed occupancy prediction pipeline. The whole pipeline is designed by transformer paradigm with initial point position, 3D Gaussian representation, query, and continuous multiview image frames. }
\label{fig}
\end{figure*}

\subsection{Overview}
The framework of our method is shown in Figure \ref{fig}. We use multi-view images $\mathcal{I}=\left\{\mathbf{I}_i \in \mathbb{R}^{3 \times H \times W} \mid i=1, \ldots, N\right\}$ as input and first utilize ResNet-50 \cite{resnet} as the backbone to encode image features $\mathbf{F}$. At the same time, we initialize a set of learnable queries $\mathbb{Q}$, point positions $\mathbb{P}$, and 3D Gaussians $\mathbb{G}$ to capture spatial structure and predict the occupied locations and semantic classes.
In the core component, the two-modal decoder including six decoder layers shares the same form as the decoder part in Transformer \cite{2017Attention} but with specific modifications. This decoder layer is composed of two branches. The first branch is the point decoder, which uses queries $q \in \mathbb{Q}$ and point positions $p \in \mathbb{P}$ to aggregate image features through consistent point sampling. After adaptive mixing and self-attention operation \cite{opus}, the query features are generated to update the properties of the 3D Gaussians $\mathcal{G} \in \mathbb{G} $ and predict point-level semantic classification. Using the Gaussian-to-voxel splatting module from \cite{gaussianformer}, local Gaussian distributions are aggregated to fuse with point-level semantics to predict 3D occupancy categories $\mathbf{O} \in \mathcal{C}^{X \times Y \times Z}$, where $\mathcal{C}$ and $\{X, Y, Z\}$ denote semantic classes and volume resolution. The predicted values at each layer of the decoder are supervised to ensure sufficient training of our end-to-end framework.

\subsection{Gaussian Properties}
In this paper, we use 3D Gaussians to represent regions of interest in space where potential objects may exist, unconstrained by the fixed positions of voxel grids. Compared to point-based queries, 3D Gaussians have multiple attributes, making them more suited for representing objects. The object-centered 3D Gaussian representation method allows for flexible exploration of regions of interest, progressively refining and modeling the fine-grained structure of the 3D scene in a sparse manner. Specifically, we initialize a set of Gaussian distributions $\mathcal{G}_i \in \mathbb{R}^d $, where each 3D Gaussian distribution is represented by a d-dimensional vector. The d dimension consists of $\mathbf{m} \in \mathbb{R}^3, \mathbf{s} \in \mathbb{R}^3, \mathbf{r} \in \mathbb{R}^4, \mathbf{c} \in \mathbb{R}^{|\mathcal{C}|}$. Here, $\mathbf{m}$, $\mathbf{s}$, $\mathbf{r}$, and $\mathbf{c}$ represent the mean, scale, rotation vector, and semantics, respectively. The mean and covariance characteristics allow 3D Gaussians to have various flexible shapes, providing a highly expressive capability for scene representation. The semantic attribute binds each Gaussian's position with its corresponding semantic label, thus eliminating the need for additional decoding from high-dimensional features to obtain semantic representations. In graphics, Gaussian attributes can also include glossiness, but for the purpose of controlling parameters, we exclude attributes that are more relevant to rendering but less related to occupancy prediction. Now, the Gaussian distribution $\mathcal{G}_i$ at any point $p=(x, y, z)$ in space can be computed as follows:
\begin{equation}
\begin{gathered}
\mathcal{G}(\mathbf{p} ; \mathbf{m}, \mathbf{s}, \mathbf{r}, \mathbf{c})=\exp \left(-\frac{1}{2}(\mathbf{p}-\mathbf{m})^T \boldsymbol{\Sigma}^{-1}(\mathbf{p}-\mathbf{m})\right) \mathbf{c} \\
\boldsymbol{\Sigma}=\mathbf{R S S}^T \mathbf{R}^T, \quad \mathbf{S}=\operatorname{diag}(\mathbf{s}), \quad \mathbf{R}=\operatorname{q}2\mathrm{r}(\mathbf{r})
\end{gathered}
\end{equation}
where $\boldsymbol{\Sigma}$, $\operatorname{diag}(\cdot)$ and $\mathrm{q}2\mathrm{r}(\cdot)$ represent the covariance matrix, diagonal matrix constructor and rotation matrix constructor, respectively.

\subsection{Two-modal Decoder}
The most important component is the multi-layer dual-modal decoder. Therefore, we will now describe the internal details of the decoder.

\begin{figure*}[t]
\centering
\includegraphics[width=0.8\linewidth]{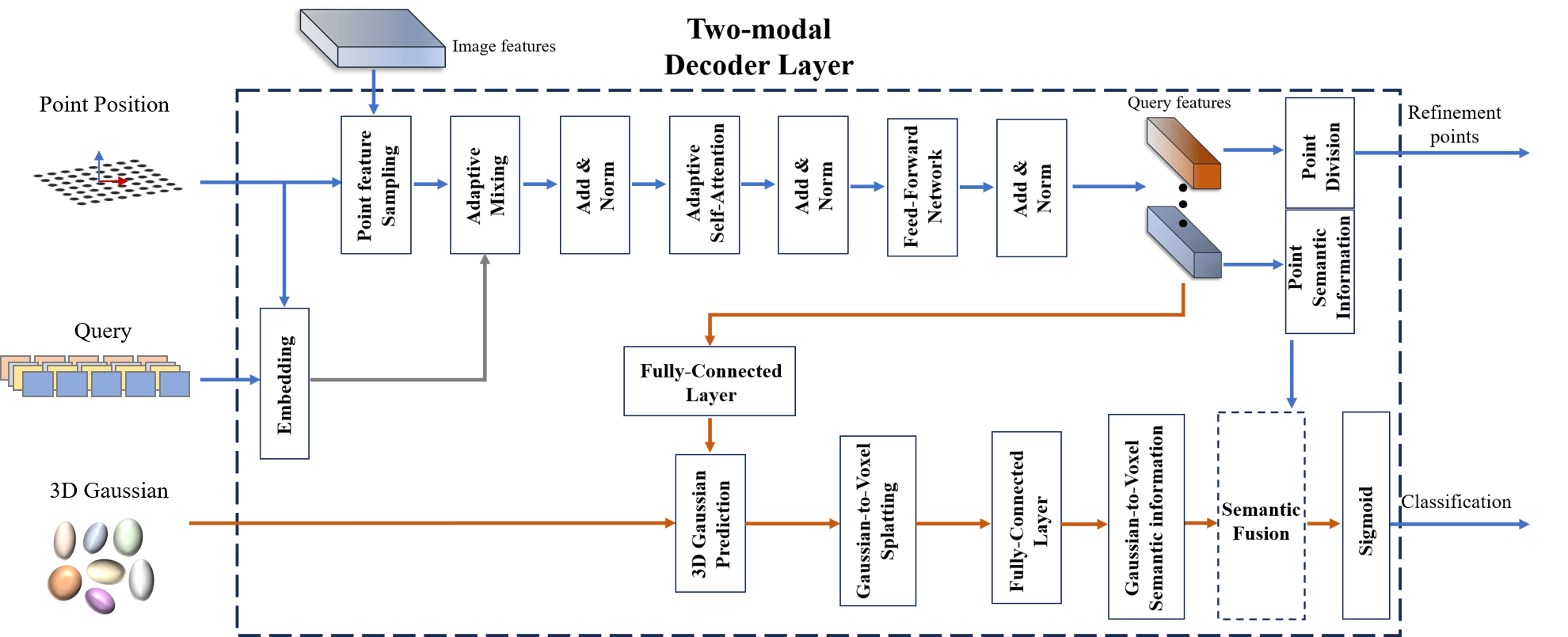}
\caption{The illustration of two-modal decoder layer. The decoder, as the core component of the pipeline, is designed in the transformer architecture to perform key functions, including image feature sampling, generation of query features through the attention mechanism, and updating of Gaussian attributes.}
\label{fig_first_case}
\end{figure*}

\textbf{Input:} In the previous section, we introduced the initial physical properties of the Gaussian distributions $\mathbb{G}_0$, which are the targets for the model to learn. However, unstructured Gaussian distributions are difficult to handle in some deformation-based attention encoding modules. Therefore, we also initialize a set of learnable 3D points $\mathbb{P}_0$, which share the same position as the 3D Gaussians, as well as high-dimensional query features $\mathbb{Q}_0$. These are used to extract image features in the consistency sampling and adaptive mixing modules, implicitly encoding 3D information, and guiding the stepwise update of the 3D Gaussians.
Thus, the input to the dual-modal decoder consists of the set $\{\mathbb{G},\mathbb{P},\mathbb{Q}\}$. The output at each layer consists of refined point positions and intermediate results of the occupancy prediction.

\textbf{Decoder Details:} To improve efficiency and save computational resources, we follow OPUS and choose a fully sparse and efficient decoding method. For a given query $q \in \mathbb{R}^{Q\times256} $ and its corresponding point position $p \in \mathbb{R}^{Q\times R \times 3}$, the decoder first performs consistent point sampling to extract image features, where $Q$ represents the set length and $R$ represents $R$ points to be predicted for each query $q$. Given the input $\{q,p\} \in \{ \mathbb{P},\mathbb{Q} \}$, we set the number of sampled points to 4 and map their respective coordinates in the $m-th$ image feature $\mathbf{c}_m$ using the following equation:
\begin{equation}
\mathbf{c}_m=\mathbf{T}_{\mathbf{m}} \mathbf{r}, \text { where } \mathbf{r}=\mathbf{m}_p+\phi(\mathbf{q}) \cdot \sigma_p
\end{equation}
Where $\mathbf{T}_{\mathbf{m}}$ represents the projection matrix from the current 3D space to the $m-th$ image coordinates. $\phi(\mathbf{q})$ uses a linear layer to generate $S$ 3D points from the query features $q$. $\mathbf{m}_p$ and $\sigma_p$ represent the mean and standard deviation of the $R$ points in $p$, respectively. The offset $\phi(\mathbf{q})$ and the standard deviation $\sigma_p$ are weighted to act as a correction mechanism, preventing the sampling process from overly focusing on more prominent points and thus causing the sampling range to become increasingly concentrated.

Subsequently, the sampled features $\mathbf{c}_m$ are adaptively mixed with the query and then passed through a self-attention layer for feature aggregation, allowing the model to focus on important information. The updated query features are then obtained through residual connections and a feed-forward network. Finally, a prediction head, composed of Linear, LayerNorm, and ReLU layers, generates the semantic class $c_i$ and position offset $\Delta p$, which are used to update the point positions, enabling the extraction of more effective image features.

The updated query features are used to guide the correction and update of the Gaussian attributes. Specifically, we employ a multi-layer perceptron (MLP) to derive intermediate attributes $\hat{G} = (\hat{m}, \hat{s}, \hat{r}, \hat{c})$ from the query features. For the intermediate mean $\hat{m}$, it is added to the original mean $m$ in a residual manner to obtain the updated attribute. The other intermediate attributes directly replace their original counterparts to become the updated values.
As described in Equation (1), Gaussian distributions at a point $p$ are aggregated by summation. 
To reduce computational complexity, we adopt the Gaussian-to-Voxel Splatting approach \cite{gaussianformer} for efficient indexing, generating point-alignment Gaussian semantic information $g_i$. After that, a semantic fusion mechanism is deployed to fuse $c_i$ and $g_i$ to generate reliable semantic classification \(\mathbf{O} \in \mathcal{C}^{X \times Y \times Z}\), where subscript $i$ denotes the $i-th$ decoder layer. As can be seen in Fig \ref{fig_first_case}, the fusion format can be formulated as:
\begin{equation}
\mathbf{O} = \frac{1}{2} \left[ \text{Sigmoid}(FCN(g_i)) \cdot c_i + \text{Sigmoid}(c_i) \cdot g_i \right]
\end{equation}
in which the $FCN$ (fully-connected layer) is used to align the feature dimension of $c_i$ and $g_i$. With this fusion operation, the semantic information for occupancy classification can be adaptively enhanced.

\subsection{Loss Functions}
Our model is efficiently trained in an end-to-end manner, using ground truth to supervise learning. Following \cite{point,self,opus}, we adopt a weighted Chamfer Distance to match the predicted points set $\mathbb{P}$ with the ground truth points $\mathbb{P}_g$. This approach not only focuses on overall accuracy but also emphasizes penalizing erroneous points. The weighted Chamfer Distance is defined as:
\begin{equation}
\begin{array}{r}
\mathrm{CD}_R\left(\mathbb{P}, \mathbb{P}_g\right)=\frac{1}{|\mathbb{P}|} \sum\limits_{\mathbf{p} \in \mathbb{P}} D_R\left(\mathbf{p}, \mathbb{P}_g\right)+\frac{1}{\left|\mathbb{P}_g\right|} \sum\limits_{\mathbf{p}_g \in \mathbb{P}_g} D_R\left(\mathbf{p}_g, \mathbb{P}\right), \\
\text { where } D_R(\mathbf{x}, \mathbb{Y})=W(d) \cdot d \text { with } d=\min _{\mathbf{y} \in \mathbb{Y}}\|\mathbf{x}-\mathbf{y}\|_1 .
\end{array}
\end{equation}

In the formula, $W(d)$ is a re-weighting function that penalizes points with large distances towards the ground truth. In the specific implementation, a step function is used to define $wd$. If $d \geq 0.2$, then $w(d)=5$; otherwise, $w(d)=1$. This weighting mechanism assigns higher penalties to points with larger errors $(d \geq 0.2)$, encouraging the model to focus on correcting significant deviations during training.

For the classification task, we follow OPUS \cite{opus} to use focal loss $L_{focal}$ to measure supervision loss. The overall loss function is defined as:
\begin{equation}
L=\mathrm{CD}_R\left(\mathbb{P}_0, \mathbb{P}_g\right)+\sum_{i=1}^6\left(
\mathrm{CD}_R\left(\mathbb{P}_i, \mathbb{P}_g\right)+ L_{focal}^i
\right)
\end{equation}

\section{Experiments}
\subsection{Experimental setup}

\begin{table*}[t]
    \caption{The performance of occupancy predcition on Occ3D-nuScenes. "8f" and "16f" means the number of input frames with 8 frames and 16 frames. \label{tab1}}
\setlength{\tabcolsep}{3pt}
    \begin{center}
    \begin{tabular}{ c| c | c | c | c  c  c  c | c }
    \hline
    Methods & Backbone & Resulution & 
    \multicolumn{1}{c|}{mIoU} & 
    \multicolumn{1}{c}{$RayIoU_{1m}$} & 
    \multicolumn{1}{c}{$RayIoU_{2m}$} & 
    \multicolumn{1}{c}{$RayIoU_{4m}$} & 
    \multicolumn{1}{c|}{$RayIoU$} &
    \multicolumn{1}{c}{FPS} \\
    \hline
    RenderOcc \cite{renderocc} & Swin-B & 1408×512 & 24.5 & 13.4 & 19.6 & 25.5 & 19.5 & -\\
    BEVFormer \cite{BEVFormer} & R101 & 1600×900 & 39.3 & 26.1 & 32.9 & 38.0 & 32.4 & 3.0 \\
    BEVDet-Occ \cite{bevdepth} & R50 & 704×256 & 36.1 & 23.6 & 30.0 & 35.1 & 29.6 & 2.6 \\
    BEVDet-Occ(8f) \cite{bevdepth} & R50 & 704×384 & 39.3 & 26.6 & 33.1 & 38.2 & 32.6 & 0.8 \\
    FB-Occ(16f) \cite{fbocc} & R50 & 704×256 & 39.1 & 26.7 & 34.1 & 39.7 & 33.5 & 10.3 \\
    Sparse-Occ(8f) \cite{sparseocc} & R50 & 704×256 & - & 28.0 & 34.7 & 39.4 & 34.0 & 17.3 \\
    Sparse-Occ(16f) \cite{sparseocc} & R50 & 704×256 & 30.6 & 29.1 & 35.8 & 40.3 & 35.1 & 12.5 \\
    OPUS-T(8f) \cite{opus} & R50 & 704×256 & 33.2 & 31.7 & 39.2 & 44.3 & 38.4 & 22.4 \\
    OPUS-S(8f) \cite{opus} & R50 & 704×256 & 34.2 & 32.6 & 39.9 & 44.7 & 39.1 & 20.7 \\

  \hline
  TGP-T(8f) & R50 & 704×256  & 33.4 & 31.8 & 39.5 & 44.6 & 38.6 & 18.6\\
  TGP-S(8f) & R50 & 704×256  & 34.5 & 32.8 & 40.2 & 45.2 & 39.4 & 15.1\\
  
  \hline
  \end{tabular}
  \end{center}
\end{table*}

\begin{figure*}[t]
\centering
\includegraphics[width=0.8\linewidth]{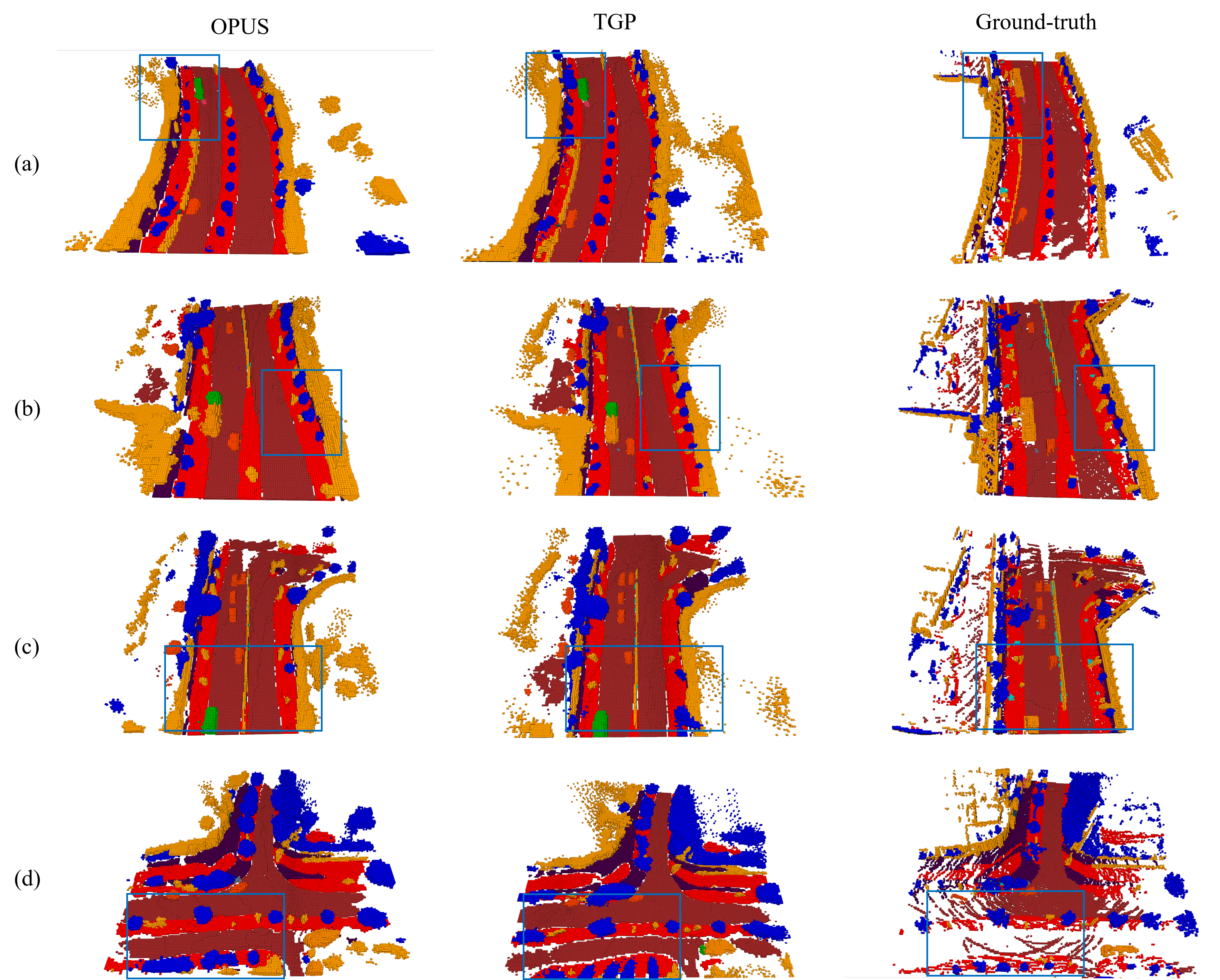}
\caption{Visualization comparison in four scenarios.}
\label{VIS}
\end{figure*}

\subsubsection{Dataset and metrics}We evaluated the proposed method on the widely recognized Occ3D-nuScenes \cite{occ3dnu} dataset, a large-scale benchmark developed to support 3D occupancy prediction. The Occ3D-nuScenes dataset comprises data from six surround-view cameras, one LiDAR, and five RaDAR, encompassing a total of 600 training scenes, 150 validation scenes, and 150 test scenes, amounting to 40,000 frames. The dataset annotations include 18 categories, consisting of one free-form category and 17 semantic categories. SparseOcc \cite{sparseocc} introduced a ray-based evaluation metric, RayIoU, to address the inconsistencies in depth penalties inherent in the traditional mIoU standard. Consequently, this study adopts the aforementioned metric and computes RayIoU at different thresholds of 1 meter, 2 meters, and 4 meters, with the final results obtained by averaging these values.

\subsubsection{Implementation details} Our algorithm is implemented using PyTorch, with ResNet-50 \cite{resnet} chosen as the image backbone, processing input images at a resolution of 256×704. We define two versions based on the number of queries, 3D Gaussians, and initial sparse points: 600 (TGP-T) and 2400 (TGP-S). The number of 3D Gaussians remains fixed at each layer, while the number of sparse points splits progressively across layers, following the same configuration used in OPUS. During training, the AdamW \cite{adam} optimizer is utilized with a learning rate of 2e-4, and a cosine annealing strategy is adopted for learning rate decay. The batch size is set to 8. All experiments are conducted with 100 epochs for validation and 24 epochs for ablation study on 4 NVIDIA A40L GPUs.

\subsection{Quantitative Comparison of Results}
The performance report for occupancy prediction on the Occ3D-nuScenes dataset is presented in Table \ref{tab1}, showcasing a comparison between our method and previous state-of-the-art approaches. As shown, our method achieves superior results across the metrics, including mIoU, RayIoU${1m}$, RayIoU${2m}$, RayIoU$_{4m}$, and RayIoU, with scores of 34.5, 32.8, 40.2, 45.2, and 39.4, respectively. Compared to similar point-based methods such as OPUS \cite{opus}, our approach delivers significantly better performance under identical configurations, albeit with a slightly lower FPS. This trade-off suggests that combining 3D Gaussian representations with sparse point-based occupancy prediction enhances accuracy while incurring some loss in inference speed. In conclusion, we introduce an innovative paradigm that integrates coexisting volumes and points, providing a valuable contribution to improving occupancy prediction accuracy and offering meaningful insights for future research.

\subsection{Qualitative Results}
We provide detailed visualization results to further demonstrate the effectiveness of our method. In Fig. \ref{VIS}, we showcase four different occupancy scenarios, each highlighting the performance of our method in comparison to OPUS \cite{opus} and the ground truth. The discrepancies between our method, OPUS, and the ground truth are marked by blue boxes for better clarity. Upon inspection, it is evident that our method aligns more closely with the ground truth, particularly in terms of volumetric occupancy. This closer alignment indicates that our approach captures the spatial structure and occupancy details more accurately, which reinforces the superiority and effectiveness of our method in real-world applications.

\begin{table}[t]
    \caption{Performance analysis of two-modal decoder layer (GS). The GP$_s$ and GP$_i$ represents initialize the 3D Gaussian with/without sharing the same value of sparse point postions. \label{tab2}}
\setlength{\tabcolsep}{3pt}
    \begin{center}
    \begin{tabular}{ c | c | c | c  c  c  c}
    \hline
    $GP_i$ & $GP_s$& 
    mIoU & 
    \multicolumn{1}{c}{$RayIoU_{1m}$} & 
    \multicolumn{1}{c}{$RayIoU_{2m}$} & 
    \multicolumn{1}{c}{$RayIoU_{4m}$} & 
    \multicolumn{1}{c}{$RayIoU$} \\
    \hline
    - & - & 32.0 & 30.1 & 37.7 & 43.0 & 36.9 \\
    \usym{1F5F8} & - & 31.9 & 30.1 & 37.8 & 43.1 & 37.0 \\
    - & \usym{1F5F8} & 32.3 & 30.2 & 38.1 & 43.4 & 37.2 \\
  
  \hline
  \end{tabular}
  \end{center}
\end{table}

\subsection{Ablation study}
To validate the effectiveness of our proposed two-modal occupancy prediction framework, which combines 3D Gaussian and sparse points, we conducted ablation studies to analyze the contribution of the two-modal decoder layer (GS). The baseline is set without 3D Gaussian modality in decoder layer as same in OPUS \cite{opus}. 
Table \ref{tab2} presents the performance improvements achieved by incorporating GS. Obviously, utilizing GS significantly enhances the performance of the occupancy prediction task when compared to the baseline, which proves the effectiveness of GS. Additionally, Table \ref{tab2} reports a comparison of different position initialization strategies for the 3D Gaussian. It can be observed that initializing the position of the 3D Gaussian without sharing the same values as the sparse points results in only slight performance gains for the occupancy prediction task. We hypothesize that this phenomenon is due to the distribution mismatch between the 3D Gaussian and sparse points, caused by the differing position initializations. Therefore, it is crucial to ensure consistency between the 3D Gaussian and sparse points at the initialization stage.

\section{Conclusion}
In this work, we innovatively propose an occupancy prediction method that adopts a dual-modal representation based on 3D Gaussian and sparse points. This approach effectively integrates the volumetric occupancy information of 3D Gaussian with the positional information of sparse points, significantly enhancing the performance of occupancy prediction. However, the introduction of 3D Gaussian has a certain impact on inference speed, which will be a key focus for improvement and optimization in future work.


\ifCLASSOPTIONcaptionsoff
  \newpage
\fi



%
\bibliographystyle{IEEEtran} 
\bibliography{ref} 


%




\end{document}